\documentclass[letterpaper]{article} % DO NOT CHANGE THIS
\usepackage{aaai2026}
\usepackage{times}  % DO NOT CHANGE THIS
\usepackage{helvet}  % DO NOT CHANGE THIS
\usepackage{courier}  % DO NOT CHANGE THIS
\usepackage[hyphens]{url}  % DO NOT CHANGE THIS
\usepackage{graphicx} % DO NOT CHANGE THIS
\usepackage{natbib}  % DO NOT CHANGE THIS AND DO NOT ADD ANY OPTIONS TO IT
\usepackage{caption} % DO NOT CHANGE THIS AND DO NOT ADD ANY OPTIONS TO IT
\usepackage{algorithm}
\usepackage{algorithmic}
\usepackage{amsfonts}

\usepackage{newfloat}
\usepackage{listings}

\usepackage{tikz}
\usepackage{booktabs}

\DeclareCaptionStyle{ruled}{labelfont=normalfont,labelsep=colon,strut=off} % DO NOT CHANGE THIS
\floatstyle{ruled}
\newfloat{listing}{tb}{lst}{}
\floatname{listing}{Listing}
\title{Attention as Binding: A Vector-Symbolic Perspective on Transformer Reasoning}
\author{
    Sahil Rajesh Dhayalkar
}
\affiliations{
    Arizona State University\\
    sdhayalk@asu.edu
}

\usepackage{bibentry}
\begin{document}

\maketitle

\begin{abstract}
Transformer-based language models display impressive reasoning-like behavior, yet remain brittle on tasks that require stable symbolic manipulation. This paper develops a unified perspective on these phenomena by interpreting self-attention and residual streams as implementing an approximate Vector Symbolic Architecture (VSA). In this view, queries and keys define role spaces, values encode fillers, attention weights perform soft unbinding, and residual connections realize superposition of many bound structures. We use this algebraic lens to relate transformer internals to chain-of-thought traces, program-based reasoning, and memory-augmented tool use, and to explain characteristic failure modes such as variable confusion and inconsistency across logically related prompts. Building on this perspective, we propose VSA-inspired architectural biases, including explicit binding/unbinding heads and hyperdimensional memory layers, and training objectives that promote role–filler separation and robust superposition. Finally, we outline metrics for measuring “VSA-likeness'' and logical compositionality, and pose theoretical and architectural open problems. Overall, the paper argues that viewing attention as soft vector-symbolic computation offers a principled route toward more interpretable and logically reliable reasoning systems.
\end{abstract}

% Uncomment the following to link to your code, datasets, an extended version or similar.
% You must keep this block between (not within) the abstract and the main body of the paper.
% \begin{links}
%     \link{Code}{https://aaai.org/example/code}
%     \link{Datasets}{https://aaai.org/example/datasets}
%     \link{Extended version}{https://aaai.org/example/extended-version}
% \end{links}

\section{Introduction}
\label{sec:intro}

\subsection{Logical and symbolic reasoning in LLMs}

Transformer-based large language models (LLMs) have become the dominant paradigm across NLP and multimodal AI~\citep{vaswani2017attention,chowdhery2022palm,devlin2019bert,brown2020language,openai2023gpt4}. Despite strong performance in in-context learning, tool use, and a wide range of reasoning-like tasks, extensive evidence shows that these capabilities do \emph{not} yield robust logical or symbolic reasoning. LLMs frequently fail under simple problem variations, struggle with systematic generalization, and exhibit logical inconsistency across related queries~\citep{creswell2022selectioninference,keysers2020measuring,lippl2025when}. Chain-of-thought (CoT) prompting improves performance on many benchmarks~\citep{wei2022chain,kojima2022zeroshot,wang2023selfconsistency}, yet models still display characteristic brittleness, overfitting to surface cues, and sensitivity to minor perturbations~\citep{creswell2022selectioninference}. These limitations suggest that transformer representations only partially align with the requirements of systematic symbolic manipulation.

A central question therefore arises: \emph{what representational operation is attention actually performing?} Attention is commonly viewed as content-addressable lookup~\citep{olsson2022incontext}, but such interpretations say little about how roles, fillers, and structured relations are internally encoded. Classical work in cognitive science and neurosymbolic AI emphasizes that robust reasoning requires mechanisms for binding and unbinding symbolic constituents~\citep{smolensky1990tensor,garcez2009neural,garcez2020neurosymbolic}. This paper adopts that perspective.

We argue that attention can be understood as a \emph{soft binding/unbinding operator}. Queries and keys define role-like subspaces; values supply fillers; and attention weights implement a differentiable unbinding step that retrieves fillers according to role similarity. Residual connections superpose many such bindings. When these operations behave coherently, transformers can approximate structured symbolic manipulation; when they do not, we obtain the logical brittleness now extensively documented.

\subsection{From symbol manipulation to Vector Symbolic Architectures}

Vector Symbolic Architectures (VSAs), or hyperdimensional computing, provide an algebraic framework for representing symbols and structures in high-dimensional vectors~\citep{plate1995hrR,kanerva2009hyperdimensional,gayler2003vsa,Kleyko2022vsa}. Atomic symbols are mapped to random high-dimensional vectors, and complex structures are built using three core operations: \emph{binding} (e.g., circular convolution, XOR, elementwise multiplication) to form role--filler pairs; \emph{superposition} (addition) to store sets or multisets of such bindings; and \emph{permutation} to encode order or hierarchy. These operations support approximate \emph{unbinding}, enabling retrieval of fillers using similarity search. 

VSAs therefore implement a distributed yet compositional form of symbolic computation~\citep{kanerva2009hyperdimensional,frady2021variablebinding,Kleyko2022vsa}. Because this algebra remains closed in a fixed vector space, it provides explicit role--filler structure and robust superposition that classical symbolic reasoning demands but standard neural embeddings rarely guarantee~\citep{garcez2009neural,garcez2020neurosymbolic}.

The central thesis of this review is that \emph{transformer attention can be interpreted as a soft, approximate instance of this VSA algebra}. Queries and keys define role-like vectors, values supply fillers, and attention weights implement a differentiable unbinding operation. Residual connections act as superposition, and multi-head structure enables layered compositions. Logical brittleness arises when these approximations fail---e.g., embeddings are insufficiently decorrelated or roles and fillers interfere---mirroring known VSA failure modes.

\subsection{Scope and contributions of this review}

Rather than a conventional survey, this paper offers a \emph{conceptual synthesis} connecting three strands of research:
\begin{enumerate}
    \item Transformer/attention mechanisms and reasoning techniques such as chain-of-thought and tool-augmented models~\citep{vaswani2017attention,wei2022chain,wang2023selfconsistency,mialon2023augmented}.
    \item Vector Symbolic Architectures as a principled algebra for compositional vector representations~\citep{plate1995hrR,kanerva2009hyperdimensional,Kleyko2022vsa}.
    \item Neurosymbolic reasoning frameworks integrating statistical and symbolic computation~\citep{garcez2009neural,garcez2020neurosymbolic,creswell2022selectioninference,mialon2023augmented}.
\end{enumerate}

Our contributions are:
\begin{itemize}
    \item A unified interpretation of attention and residual streams as performing \emph{approximate VSA-style binding, unbinding, and superposition}.
    \item A taxonomy distinguishing ``VSA-like'' from ``non–VSA-like'' transformer mechanisms.
    \item A conceptual framework linking VSA structure to chain-of-thought behavior, tool use, and logical consistency.
    \item A research agenda---including evaluation protocols and architectural proposals such as explicit binding heads and hyperdimensional memories---for building \emph{VSA-inspired reasoning architectures}.
\end{itemize}

% \begin{figure}
%     \centering
%     \includegraphics[width=0.7\linewidth]{8gwifi.org-tikz-diagram-1764894944119.png}
%     \caption{High-level roadmap of the paper}
%     \label{fig:overview}
% \end{figure}

Grounding transformer reasoning in vector-symbolic computation enables a more principled account of when LLMs can support reliable symbolic manipulation, moving beyond descriptive benchmarking toward a structured algebraic understanding of their reasoning capabilities. 
% Figure~\ref{fig:overview} summarizes how the paper connects transformer attention, vector symbolic architectures, and logical reasoning.

\section{Background: Transformers, Attention, and Logical Reasoning}
\label{sec:background}

\subsection{Transformers and scaled dot-product attention}
\label{subsec:transformers}

The transformer architecture~\citep{vaswani2017attention} processes a sequence $\mathbf{x}_{1:n}$ by embedding tokens, adding positional encodings, and applying layers of multi-head self-attention and feed-forward blocks, each wrapped in residual connections and layer normalization~\citep{ba2016layernorm}. For a single head with input $X \in \mathbb{R}^{n \times d}$,
\begin{equation}
    Q = X W_Q,\quad K = X W_K,\quad V = X W_V,
\end{equation}
and scaled dot-product attention is
\begin{equation}
    \mathrm{Attn}(Q,K,V) = \mathrm{softmax}\!\left(\frac{QK^\top}{\sqrt{d_k}}\right)V.
\end{equation}
Multi-head attention concatenates several such heads and projects them back into the residual stream, which accumulates information across layers and effectively \emph{superposes} representations generated by different tokens and heads~\citep{elhage2021interpretability}.

Operationally, attention is a form of \emph{content-addressed read/write}. The $QK^\top$ term dynamically determines which tokens influence each other, producing an input-dependent computation graph rather than a fixed routing structure~\citep{elhage2021interpretability}. This perspective already suggests that attention enforces a consistent algebra for combining and propagating representations, motivating the binding–unbinding interpretation developed in later sections.

\subsection{Key--value memory in transformers}
\label{subsec:kv-memory}

A complementary view treats attention as a differentiable key–value store~\citep{graves2014ntm,weston2015memnn}. Each row of $K$ acts as an address and the corresponding row of $V$ as content; a query $q_i$ retrieves a similarity-weighted combination of values. In decoder-only models, \emph{KV caching} extends this mechanism across timesteps, turning the transformer into a recurrent memory system that appends new key–value pairs during generation~\citep{brown2020language,liu2023lostintheMiddle}.

Each attention head forms a distinct memory channel, projecting the residual stream into different key and value subspaces. Interpreted symbolically, keys can encode \emph{roles} or \emph{variables} (e.g., subject vs.\ object, premise vs.\ hypothesis), while values encode the corresponding \emph{fillers}. The learned matrices $W_Q,W_K,W_V$ decide which components of the residual stream participate in addressing and which become memory contents. This memory-centric interpretation provides an initial bridge from attention to symbolic variable binding and memory management~\citep{smolensky1990tensor,graves2016dnc,elhage2021interpretability}.

\subsection{Logical and symbolic reasoning in LLMs}
\label{subsec:logical-llms}

Pretraining and instruction-tuning produce surprisingly strong reasoning-like behaviors~\citep{brown2020language,chowdhery2022palm,openai2023gpt4}. Chain-of-thought prompting~\citep{wei2022chain} improves arithmetic, commonsense, and symbolic tasks~\citep{cobbe2021gsm8k,kojima2022zeroshot,nye2021showyourwork}, and self-consistency~\citep{wang2023selfconsistency} further boosts performance. Separately, LLMs augmented with external tools---code interpreters~\citep{chen2021codex,gao2023pal,schick2023toolformer}, theorem provers~\citep{polu2020generative,wu2022autoformalization}, and logical/probabilistic solvers~\citep{pan-etal-2023-logic,creswell2022selectioninference,mialon2023augmented}---can solve more complex problems by delegating computation.

Yet persistent limitations remain. LLMs frequently exhibit logical inconsistency across related prompts~\citep{razeghi2022consistency,jin-etal-2025-disentangling-memory}, struggle with systematic generalization and variable substitution~\citep{keysers2020measuring,hupkes2020compositional}, rely on superficial cues rather than underlying structure~\citep{min2022rethinking}, and are sensitive to small perturbations in problem phrasing~\citep{perez2021trueReasoning}. These issues indicate that transformers can \emph{approximate} symbolic reasoning but do not inherently guarantee reliable symbolic computation.

\subsection{Why algebraic views of attention matter for logic}
\label{subsec:algebraic-view}

Logical reasoning requires \emph{consistent manipulation of structured representations}: variables must be bound and renamed without collision, predicates must compose systematically, and inference rules must preserve truth across transformations~\citep{enderton2001logic,mitchell1996foundations,garcez2020neurosymbolic}. Neural architectures aiming to support such reasoning must therefore implement disciplined operations analogous to binding, unbinding, and superposition.

Standard interpretations of attention---as token-weight distributions or dynamic feature selectors---do not specify how roles and fillers are represented or combined, making it difficult to articulate or constrain the invariances logic demands. This motivates an explicitly \emph{algebraic} framework for understanding attention.

Vector Symbolic Architectures offer such a framework: a small set of operations (binding, superposition, permutation) with well-defined algebraic properties for symbolic computation in high-dimensional spaces~\citep{plate1995hrR,kanerva2009hyperdimensional,Kleyko2022vsa}. Recasting attention through this lens clarifies when its behavior approximates principled binding/unbinding and when it departs from it, illuminating both strengths and limitations of LLMs as logical reasoners.

\section{Vector Symbolic Architectures: Algebra for Compositional Vectors}
\label{sec:vsa}

\subsection{Core principles of VSAs}
\label{subsec:vsa-principles}

Vector Symbolic Architectures (VSAs), or hyperdimensional computing, represent discrete symbols by dense, high-dimensional vectors that are approximately orthogonal~\citep{plate1995hrR,kanerva2009hyperdimensional,gayler2003vsa,Kleyko2022vsa}. Let $\mathcal{H} \subset \mathbb{R}^D$ denote this space. Symbolic structure is encoded using three algebraic operations:
\begin{itemize}
    \item \textbf{Binding} ($\mathbf{a}\otimes\mathbf{b}$) forms role--filler pairs using elementwise multiplication, circular convolution, or XOR~\citep{plate1995hrR,rachkovskij2001binary}.
    \item \textbf{Superposition} sums vectors to represent sets or multisets~\citep{kanerva2009hyperdimensional}.
    \item \textbf{Permutation} applies fixed invertible transformations to encode order or structure~\citep{plate1995hrR}.
\end{itemize}
These operations remain within $\mathcal{H}$ and support approximate \emph{unbinding}: given a bound vector and one constituent, the other can be recovered via similarity search. This closed algebra enables a form of symbolic computation directly in vector space~\citep{gayler2003vsa,Kleyko2022vsa}.

\subsection{Representing structures: sequences, sets, trees, graphs}
\label{subsec:vsa-structures}

Complex structures arise from repeated binding, superposition, and permutation. A sequence $(x_1,\dots,x_n)$ may be encoded as
\begin{equation}
    \mathbf{s} = \sum_{i=1}^n \pi^i(\mathbf{pos}) \otimes \mathbf{x}_i,
\end{equation}
retrieved by unbinding with $\pi^i(\mathbf{pos})$~\citep{plate1995hrR,kanerva2009hyperdimensional}. Sets correspond to pure superpositions, while key--value stores take the form $\sum_j \mathbf{k}_j \otimes \mathbf{v}_j$. Trees and graphs follow similar principles: subtrees or edges are bound to role vectors such as \textsc{left}, \textsc{right}, or edge labels and then superposed~\citep{frady2021variablebinding,Kleyko2022vsa}. All representations remain in the same vector space, with decoding performed via similarity, giving VSAs a uniform representational substrate for structured data.

\subsection{Neurosymbolic reasoning with VSAs}
\label{subsec:vsa-neurosymbolic}

Since VSA operations support similarity-based retrieval and unbinding, they naturally implement associative memories, symbolic rule application, and multi-step inference~\citep{kanerva2009hyperdimensional,frady2021variablebinding,Kleyko2022vsa}. Predicates, for instance, can be represented as vectors $\mathbf{p}$ such that $\mathbf{p}\otimes\mathbf{x}$ encodes ``$P(x)$,'' with logical combinations handled via superposition. These properties make VSAs attractive for neurosymbolic models of working memory, compositional semantics, and cognitive reasoning~\citep{plate1995hrR,gayler2003vsa,kanerva2009hyperdimensional}. Moreover, binding and superposition are differentiable or readily approximated, enabling neural architectures that manipulate VSA representations while preserving a symbolic interpretation.

%%%%%%%%%%%%%%%%%%%% commented because I am using a table
% \subsection{Comparison with other compositional frameworks}
% \label{subsec:vsa-comparison}

% VSAs sit among several frameworks for distributed compositional representations. Tensor Product Representations (TPRs)~\citep{smolensky1990tensor} bind roles and fillers via higher-order tensors, while Holographic Reduced Representations (HRRs)~\citep{plate1995hrR} use circular convolution as a binding operator. Structured neural networks, graph neural networks~\citep{scarselli2009gnn,kipf2017gnn}, and neural program-induction architectures~\citep{andreas2016neural,graves2016dnc} encode structure via explicit graphs or pointer-based memories.

% VSAs occupy a middle ground: like TPRs, they provide explicit role–filler separation; unlike TPRs, they maintain fixed-dimensional vectors, aligning naturally with transformer embeddings and residual streams. Unlike graph-based approaches, VSAs do not require explicit symbolic data structures—the structure is implicit in the algebra. This combination of algebraic clarity, fixed dimensionality, and compatibility with similarity-based decoding makes VSAs especially well-suited as a conceptual bridge to transformers, motivating our reinterpretation of attention as a soft VSA binding and unbinding operator.

\begin{table*}[!t]   % <-- note the *
\centering
\small
\begin{tabular}{p{2.0cm}p{3.5cm}p{4.75cm}p{5.0cm}}
\toprule
\textbf{Property} &
\textbf{VSAs / HRRs} &
\textbf{Transformers (this paper)} &
\textbf{TPRs / GNNs / others} \\
\midrule
Base representational object &
Fixed-dimensional high-dimensional vectors in $\mathcal{H}$ &
Token embeddings / residual stream vectors in fixed width &
Higher-order tensors (TPRs) or graph nodes/edges (GNNs, structured nets) \\
\midrule
Binding mechanism &
Algebraic operator $\otimes$ (e.g., XOR, convolution, elementwise product)~\citep{plate1995hrR,kanerva2009hyperdimensional} &
Soft attention over keys and values; queries act as role cues, values as fillers &
Outer product between role/filler vectors (TPRs)~\citep{smolensky1990tensor}; message passing or edge updates in GNNs~\citep{scarselli2009gnn,kipf2017gnn} \\
\midrule
Superposition / storage of sets &
Vector addition of bound items, often with normalization~\citep{kanerva2009hyperdimensional} &
Residual connections accumulate outputs of attention and MLPs into a running superposition &
Explicit sets or graphs; sometimes implicit via summation / pooling over nodes or features \\
\midrule
Positional / structural encoding &
Fixed permutations or rotations $\pi$ applied to role/filler vectors~\citep{plate1995hrR} &
Positional encodings and RoPE/ALiBi; can be viewed as differentiable permutations acting prior to attention &
Graph topology or program structure encodes position; sometimes positional embeddings or pointer structures \\
\midrule
Decoding / unbinding &
Similarity search (e.g., cosine) to recover fillers from bound/superposed vectors &
Attention and probes perform approximate unbinding from hidden states; similarity-based decoders for analysis &
Tensor projection (TPRs) or graph algorithms (message passing, readout) extract structured information \\
\midrule
Advantages for symbolic reasoning &
Closed algebra; explicit role--filler binding; robust superposition; easy similarity-based retrieval &
Integrates VSA-like behavior into scalable sequence models; compatible with large-scale pretraining and tool use &
Very explicit structure (TPRs, graphs) and strong inductive bias for relational reasoning; often highly interpretable \\
\midrule
Limitations / challenges &
Approximate unbinding; capacity--interference trade-offs; usually hand-designed operators &
VSA-like behavior only approximate; interference, dense attention, and entangled roles/fillers can break symbolic discipline &
Higher-order tensors expensive; graph/program structure may be hard to induce from raw data or scale to very long contexts \\
\bottomrule
\end{tabular}
\caption{Comparison between VSAs, transformers under the VSA interpretation developed in this paper, and other compositional representation frameworks such as Tensor Product Representations (TPRs) and graph neural networks (GNNs).}
\label{tab:vsa-comparison}
\end{table*}

\subsection{Comparison with other compositional frameworks}
\label{subsec:vsa-comparison}

VSAs sit among several frameworks for distributed compositional representations. Tensor Product Representations (TPRs)~\citep{smolensky1990tensor} bind roles and fillers via higher-order tensors, while Holographic Reduced Representations (HRRs)~\citep{plate1995hrR} instantiate a particular VSA using circular convolution. Structured neural networks, graph neural networks~\citep{scarselli2009gnn,kipf2017gnn}, and neural program-induction architectures~\citep{andreas2016neural,graves2016dnc} instead encode structure in explicit graphs or pointer-based memories. As summarized in Table~\ref{tab:vsa-comparison}, VSAs occupy a middle ground: like TPRs, they support explicit role–filler binding, but they retain fixed-dimensional vectors that align naturally with transformer embeddings and residual streams; like graph-based approaches, they facilitate relational reasoning, but without requiring an explicit symbolic graph datastructure. This combination of algebraic clarity, fixed dimensionality, and compatibility with similarity-based decoding makes VSAs a particularly apt conceptual bridge to transformers, motivating our reinterpretation of attention as a soft VSA binding and unbinding operator.

\section{Attention as soft binding and unbinding}
\label{sec:attn-as-binding}

\subsection{Dot-product attention as similarity-based addressing}
\label{subsec:dotprod-similarity}

Scaled dot-product attention for a single head computes
the attention as mentioned in equation (2),
% \begin{equation}
    % \mathrm{Attn}(Q,K,V)=\mathrm{softmax}\!\left(\frac{QK^\top}{\sqrt{d_k}}\right)V,
% \end{equation}
where $Q,K,V \in \mathbb{R}^{n\times d_k}$ are derived from the residual stream~\citep{vaswani2017attention}. For position $i$,
\begin{equation}
% \[
    \mathbf{o}_i=\sum_j \alpha_{ij}\mathbf{v}_j,\quad
    \alpha_{ij}=\frac{\exp(\langle \mathbf{q}_i,\mathbf{k}_j\rangle/\sqrt{d_k})}{\sum_{j'}\exp(\langle\mathbf{q}_i,\mathbf{k}_{j'}\rangle/\sqrt{d_k})}.
% \]
\end{equation}

The $Q,K,V \in \mathbb{R}^{n\times d_k}$ are derived from the residual stream~\citep{vaswani2017attention} when computing the scaled dot-product attention for a single head. Each query induces a similarity distribution over keys, producing a convex combination of values---a classic \emph{content-addressable memory} operation~\citep{graves2014ntm,weston2015memnn,elhage2021interpretability}.  

In VSA terms, unbinding retrieves a filler by comparing a role cue with stored bindings~\citep{plate1995hrR,kanerva2009hyperdimensional,Kleyko2022vsa}. Dot-product attention is a differentiable analogue: queries act as roles, keys as stored roles, and values as fillers, with the softmax functioning as a smooth unbinding operator.

\subsection{Keys and values as role--filler pairs}
\label{subsec:keys-values}

This suggests interpreting key and value projections as implementing a role–filler decomposition~\citep{smolensky1990tensor,elhage2021interpretability}. The key projection $W_K$ extracts the role a token plays (e.g., subject/object, premise/conclusion), while $W_V$ encodes the corresponding filler. $W_Q$ determines which roles each position seeks to access. Multi-head attention provides multiple parallel binding channels, each with its own $(W_Q^{(h)},W_K^{(h)},W_V^{(h)})$. Empirical work shows head specialization: some track syntax, others lexical or positional features~\citep{clark2019electraattention,elhage2021interpretability,rogers2020primer}. In VSA language, each head implements a distinct binding scheme in a shared high-dimensional substrate.

\subsection{Attention weights as soft binding coefficients}
\label{subsec:soft-binding}

Classical VSAs perform \emph{hard} binding, e.g.
% \begin{equation}
    % \mathbf{z}=\mathbf{r}\otimes\mathbf{f},
% \end{equation}
$\mathbf{z}=\mathbf{r}\otimes\mathbf{f},$
with unbinding recovering $\mathbf{f}$ from $\mathbf{z}$ using $\mathbf{r}$~\citep{plate1995hrR}. Transformers instead realize a \emph{soft} version: keys serve as roles, values as fillers, and $\alpha_{ij}$ measures the binding strength between a query role and stored roles. The resulting output is a weighted superposition of fillers.

Attention best approximates VSA binding when (i) role vectors (keys) are near-orthogonal to reduce interference~\citep{kanerva2009hyperdimensional,frady2021variablebinding}, (ii) attention is relatively sparse, yielding crisp role–filler matches~\citep{michel2019sixteenheads,voita2019analyzing}, and (iii) normalization (like layer norm) ensures consistent geometric structure. Under these conditions, attention functions as an approximate VSA binding/unbinding operator.

\subsection{Residual streams as superposition}
\label{subsec:residual-superposition}

Residual connections naturally implement VSA-style \emph{superposition}. With layer update
\begin{equation}
    \mathbf{x}^{(\ell+1)}=\mathbf{x}^{(\ell)}+\mathrm{Attn}^{(\ell)}(\mathbf{x}^{(\ell)})+\mathrm{MLP}^{(\ell)}(\cdot),
\end{equation}
the residual stream accumulates contributions from many heads and transformations. Each attention output corresponds to one or more bound pairs, and the residual stores their superposition over depth. Layer norm maintains useful geometry, while MLPs act as learned, nonlinear re-encodings~\citep{ba2016layernorm,elhage2021interpretability}. Together, these operations approximate a pipeline of soft binding, superposition, and rewriting within a fixed-dimensional space.

\subsection{Approximation gap: when attention fails to be VSA-like}
\label{subsec:approx-gap}

In real models, this interpretation is only approximate. Learned embeddings and key projections are not perfectly orthogonal; heads often encode multiple functions simultaneously~\citep{elhage2021interpretability,rogers2020primer}, causing \emph{interference} and imperfect unbinding. Attention patterns may be dense~\citep{michel2019sixteenheads}, blurring distinct bindings. Cross-layer context mixing further entangles roles and fillers, obscuring what is bound to what. These effects motivate defining a layer or head's \emph{VSA-likeness}: how closely its operations approximate disciplined binding and unbinding over decorrelated role–filler subspaces with controlled superposition. Later sections use this notion to design metrics, probes, and architectures that make transformers more VSA-like---and, consequently, more reliable symbolic reasoners.

\section{A taxonomy of VSA-like mechanisms in modern attention architectures}
\label{sec:taxonomy}

\subsection{Standard transformers and their variants}
\label{subsec:standard-transformers}

Many transformer variants implicitly alter the binding–unbinding behavior of attention. Sparse and structured mechanisms such as Sparse Transformers~\citep{child2019sparse}, Longformer~\citep{beltagy2020longformer}, and BigBird~\citep{zaheer2020bigbird} restrict attention to local or selected tokens, producing more \emph{focused} bindings and reducing interference in the resulting superpositions.

Positional schemes such as RoPE~\citep{su2021roformer} and ALiBi~\citep{press2022alibi} reshape representational geometry in ways closely aligned with VSA positional permutations~\citep{plate1995hrR}. RoPE's rotations can be viewed as differentiable permutation operations that compose naturally with binding and superposition.

Recurrent and memory-augmented models, including Transformer-XL~\citep{dai2019transformerxl}, Compressive Transformers~\citep{rae2020compressive}, and long-range KV caching, append historical key–value pairs over time. This creates a growing superposed store of bound pairs, whose quality depends on decisions about retention, compression, and decay, directly affecting how VSA-like the memory dynamics become.

\subsection{Explicit binding mechanisms in neural networks}
\label{subsec:explicit-binding}

Several architectures implement explicit binding and slotting operations. Memory Networks~\citep{weston2015memnn}, Neural Turing Machines, and the Differentiable Neural Computer~\citep{graves2014ntm,graves2016dnc} maintain addressable memory with differentiable read/write operations. Viewed through a VSA lens, memory cells hold bound role–filler vectors, reads perform unbinding, and writes superpose or overwrite bindings.

Slot Attention~\citep{locatello2020slotattention} learns a small set of ``slot'' vectors that act as roles binding to object features via attention. Fast-weights models~\citep{schmidhuber1992learning,ba2016fastweights} and multiplicative RNNs~\citep{sutskever2011multiplicativernn} implement dynamic parameter matrices functioning as implicit high-dimensional binding stores.

Together, these architectures already instantiate VSA-like primitives: similarity-based addressing, additive superposition, and multiplicative binding. Their primary divergence from classical VSAs lies in how explicitly the algebra of roles and fillers is enforced.

\subsection{Neuro-symbolic and logic-oriented transformers}
\label{subsec:neurosymbolic-transformers}

Neurosymbolic models integrate symbolic rules or reasoning procedures with transformer architectures. Systems such as Neural Logic Machines~\citep{dong2019nlm}, Neural Theorem Provers~\citep{rocktaschel2017ntp}, Logic Tensor Networks~\citep{serafini2016ltn}, and DeepProbLog~\citep{manhaeve2018deepproblog} combine differentiable representations with logical constraints. Recent work links LLMs with external solvers and proof engines~\citep{polu2020generative,pan-etal-2023-logic,creswell2022selectioninference,mialon2023augmented}.

These architectures vary in how VSA-like their internal operations are. Some treat transformers as black-box encoders, offloading structure to external modules. Others align attention heads or memory representations with logical roles (predicate, argument, quantifier), resulting in internal states resembling superposed sets of symbolic facts. The closer the architecture encodes roles and fillers directly in vector operations, the more naturally it fits a VSA algebraic interpretation.

\subsection{Hyperdimensional and VSA-inspired deep architectures}
\label{subsec:vsa-inspired-architectures}

A growing line of work incorporates hyperdimensional computing directly into neural models. VSA-based representations have been used as classifiers or associative memories~\citep{kanerva2009hyperdimensional}, and more recent architectures embed binding and superposition into recurrent or convolutional networks~\citep{frady2021variablebinding,Kleyko2022vsa}. Some approaches replace or augment embeddings with hyperdimensional encodings, enforcing role–filler separation and robust superposition by design.

These systems demonstrate that high-dimensional binding operations are compatible with gradient-based learning and scalable models. They validate the feasibility of incorporating VSA principles into modern deep architectures---evidence we build on in later sections when proposing VSA-inspired transformer designs aimed at improving logical and symbolic reasoning capabilities.

\section{Chain-of-thought, program traces, and VSA representations}
\label{sec:cot-vsa}

\subsection{Interpreting CoT as explicit symbolic traces}
\label{subsec:cot-as-trace}

Chain-of-thought (CoT) prompting elicits intermediate reasoning steps, effectively externalizing an \emph{execution trace} in natural language~\citep{wei2022chain,kojima2022zeroshot}. Such traces correlate strongly with improved arithmetic, commonsense, and symbolic reasoning performance~\citep{wei2022chain,wang2023selfconsistency}, and their benefits scale with model size. Conceptually, a CoT response reflects a sequence of symbolic state updates.

Evidence suggests that internal representations \emph{partially} track these decompositions even without supervision: different subtasks appear across different layers~\citep{wei2022chain}, and some hidden states encode information predictive of CoT success before reasoning tokens are produced~\citep{knowingBeforeSaying2025}. However, mismatches between generated CoT and actual computation also arise, yielding fluent but unfaithful reasoning~\citep{turpin2023cotfaithfulness}. This motivates a representational framework capable of describing both faithful and spurious traces in a unified algebraic language.

\subsection{Mapping CoT steps to VSA structures}
\label{subsec:cot-to-vsa}

From a VSA perspective, each reasoning step corresponds to adding or modifying bound role–filler pairs in the residual stream. Let $\mathbf{s}^{(t)}$ denote the internal state after $t$ steps. A VSA-like update can be written as
\begin{equation}
    \mathbf{s}^{(t+1)}=\mathbf{s}^{(t)}\oplus\sum_k \mathbf{r}_k^{(t)}\otimes\mathbf{f}_k^{(t)},
\end{equation}
where $\mathbf{r}_k^{(t)}$ encode roles (e.g., subgoal, partial result) and $\mathbf{f}_k^{(t)}$ encode fillers. Later steps may update or permute roles to reflect changing structure.

Reasoning sequences thus map to nested bindings and permutations: multi-step algebraic derivations correspond to progressively binding terms, while permutations encode ordering of intermediate quantities. CoT text can therefore be viewed as an externalization of a trajectory through a VSA-structured internal space. This view suggests concrete probes: decoding approximate role/filler vectors from hidden states and checking whether adjacent layers follow VSA-predicted structured updates.

\subsection{Program induction and execution as VSA algebra}
\label{subsec:program-vsa}

Program-based reasoning extends CoT by expressing steps as executable code. Program-of-Thoughts (PoT) prompting delegates computation to external executors~\citep{chen2022programOfThought}, and program-based CoT uses code-plus-tests as verifiable reasoning traces~\citep{yang2025codeThinkSurvey,gao2023pal}. These traces correspond to explicit program states (variables, memory, control flow).

VSAs naturally encode such states. A program environment can be written as
\begin{equation}
    \mathbf{e}=\sum_{v\in\mathcal{V}} \mathbf{var}_v\otimes\mathbf{val}_v,
\end{equation}
with additional bindings and permutations encoding control flow, call stacks, or proof-tree positions. Attention performs unbinding (reading a subgoal or variable), while MLP and subsequent layers rewrite fillers and rebind them. Under this view, both program synthesis and execution correspond to sequences of VSA operations guided by transformer parameters, regardless of whether explicit code is shown to the model.

\subsection{Implications for logical consistency and proof-like reasoning}
\label{subsec:consistency-vsa}

If CoT and program traces correspond to trajectories in a VSA-like state space, robust reasoning requires stable, interpretable role–filler structure. When roles and fillers remain separable and superposition interference is controlled, transformers can maintain coherent representations of partial proofs, variable assignments, and constraints. This should yield greater logical consistency, fewer variable-substitution errors, and more robustness to paraphrase.

This perspective enables testable hypotheses: (i) define proxies for VSA-likeness (e.g., role–filler separability, interference sensitivity), and (ii) correlate them with reasoning consistency and proof quality. Conversely, architectural or training interventions that promote VSA-like structure---clean binding, reliable unbinding, controlled superposition---should improve generalization on CoT and program-based reasoning benchmarks. Later sections build on this agenda to develop probing methods and VSA-inspired reasoning architectures.

\section{Designing VSA-inspired transformer layers for logical reasoning}
\label{sec:vsa-inspired-design}

\subsection{Inductive biases from VSA algebra}
\label{subsec:vsa-biases}

The VSA perspective highlights representational properties essential for logical reasoning: (i) \emph{role--filler separation}, keeping variables, argument positions, and contexts distinct from content; (ii) \emph{robust superposition}, enabling multiple bindings to coexist with controlled interference; and (iii) \emph{near-orthogonality} among role vectors to support reliable unbinding~\citep{plate1995hrR,kanerva2009hyperdimensional,Kleyko2022vsa,frady2021variablebinding}. These can be encouraged through architectural and training biases. Certain heads or subspaces can be designated as role or value channels, with constraints such as coherence or orthogonality. Optimization-time regularizers can further maintain orthogonal key/query subspaces and well-conditioned value projections, nudging transformers toward VSA-like binding and unbinding by design rather than emergence.

\subsection{Explicit binding/unbinding heads}
\label{subsec:explicit-vsa-heads}

A concrete proposal is to introduce specialized \emph{binding/unbinding heads} whose parameters approximate VSA operators. Some heads may perform multiplicative binding via elementwise interactions between learned role vectors and filler vectors from the residual stream, replacing standard linear value projections. Positional structure can be handled by tying projections to rotary or permutation-like matrices, mirroring VSA positional permutations~\citep{plate1995hrR,su2021roformer}. Unbinding heads may approximate inverse operators (such as correlation for convolution-based binding), with queries constructed as role cues. These modules coexist with standard attention heads and feed-forward blocks, giving the model both unconstrained expressivity and disciplined VSA-style computation. Training dynamics then determine the degree of reliance on these specialized heads while structural constraints keep them close to interpretable algebraic forms.

\subsection{Hyperdimensional memory layers for LLMs}
\label{subsec:vsa-memory}

Another direction is to incorporate \emph{hyperdimensional memory layers} implementing explicit VSA-style associative stores. Such a module maintains a memory vector $\mathbf{m}\in\mathbb{R}^D$ that accumulates formulas, partial proofs, or tool states via superposed bindings:
\begin{equation}
    \mathbf{m} \leftarrow \mathbf{m} \;\oplus\; \sum_{k} \mathbf{r}_k \otimes \mathbf{f}_k.
\end{equation}
Attention or explicit VSA operators read from this memory by unbinding with role cues, while writes bind new fillers to roles. Memory may persist across layers or timesteps, functioning as a hyperdimensional workspace for storing logical constraints and intermediate results. In tool-augmented settings~\citep{pan-etal-2023-logic,creswell2022selectioninference,mialon2023augmented}, encoders/decoders can map symbolic formulas into and out of this shared memory, allowing solvers and LLMs to operate over compatible VSA representations.

\subsection{Training objectives and regularizers}
\label{subsec:vsa-objectives}

Training objectives can explicitly promote VSA-like behavior. Orthogonality constraints on role vectors and key/query projections, via spectral penalties, Bjorck normalization, or coherence regularization, reduce interference and preserve clean role–filler structure. Auxiliary reconstruction tasks can train the model to unbind fillers from synthetic bound or superposed vectors, or to recover constituents after injected superposition noise. Logic-oriented auxiliary tasks further encourage systematic generalization, such as permuted-variable reasoning, symbolic substitution, or synthetic theorem-proving tasks where identical proof patterns must generalize across renamings~\citep{keysers2020measuring,hupkes2020compositional}. Collectively, these objectives align the geometry of attention and residual streams with VSA algebra, increasing the likelihood that LLMs learn stable, compositional reasoning mechanisms.

\section{Evaluation: measuring VSA-likeness and logical compositionality}
\label{sec:evaluation}

\subsection{Defining ``VSA-likeness'' metrics}
\label{subsec:vsa-metrics}

The VSA view suggests that some attention heads approximate principled binding/unbinding while others behave as generic feature mixers. To quantify this, we define empirical \emph{VSA-likeness} metrics.

First, \emph{role--filler recoverability} tests whether putative role and filler subspaces are separable. Using synthetic role vectors $\{\mathbf{r}_k\}$ and fillers $\{\mathbf{f}_k\}$ injected via prompts or residual edits, we measure how well a probe recovers $\mathbf{f}_k$ when cued with $\mathbf{r}_k$, analogous to VSA unbinding~\citep{plate1995hrR,kanerva2009hyperdimensional,Kleyko2022vsa}. High recoverability indicates a clean role–filler factorization.

Second, \emph{interference under superposition} measures how well unbinding works when multiple bindings are combined, e.g.\ $\mathbf{z}=\mathbf{r}_1\otimes\mathbf{f}_1\oplus\mathbf{r}_2\otimes\mathbf{f}_2$. Varying the number and similarity of bindings yields capacity and interference curves, paralleling VSA associative memory analyses~\citep{kanerva2009hyperdimensional,frady2021variablebinding}.

Third, \emph{alignment with VSA operators} assesses whether learned transformations approximate known binding/permutation operations. By fitting simple VSA models (e.g., convolution, fixed permutations) to the action of a head, we can quantify how closely its behavior adheres to a VSA algebra.

\subsection{Behavioral benchmarks for VSA-style reasoning}
\label{subsec:vsa-benchmarks}

External benchmarks should stress abilities expected from VSA-like representations. Tasks testing \emph{variable binding and systematic generalization}, such as SCAN~\citep{lake2018scan}, compositional benchmarks~\citep{keysers2020measuring,hupkes2020compositional}, and algebraic manipulation under variable renaming, evaluate whether models manipulate abstract role–filler structure rather than memorizing patterns.

A second class focuses on \emph{structured symbolic manipulation}: algebraic simplification, equation solving, and rewriting tasks~\citep{saxton2019analysing,polu2020generative}, where variation in depth and breadth tests compositional generalization. Classical logic puzzles and paraphrased inference problems assess \emph{logical invariance} across syntactically diverse but equivalent formulations~\citep{razeghi2022consistency}.

To distinguish genuine compositionality from pattern matching, benchmarks should hold out combinations of roles/fillers during training, test permutation of symbol identities, and enforce generalization to unseen rule instantiations. VSA-like models are predicted to perform more robustly under such shifts.

\subsection{Empirical probes into pretrained LLMs}
\label{subsec:vsa-probes}

These metrics motivate a probing agenda for pretrained LLMs. Representational similarity analysis (RSA)~\citep{kriegeskorte2008rsa} can compare hidden-state geometry to synthetic VSA encodings of symbolic structures, revealing whether layers approximate VSA embeddings. Linear and nonlinear probes~\citep{alain2017understanding,belinkov2017probe,hewitt2019structural} can decode roles, fillers, and bindings from hidden states, estimating recoverability.

Interventional studies edit embeddings or residual streams to inject synthetic bindings (e.g., swapping variable roles) and check whether model predictions reflect consistent unbinding and rebinding. Combined with causal attention analyses, such interventions identify heads or layers exhibiting latent VSA-like structure.

\subsection{Failure modes and diagnostic patterns}
\label{subsec:vsa-failures}

A VSA-centric view helps systematize reasoning failures as breakdowns of binding/unbinding. \emph{Variable confusion} (duplicated or swapped roles) indicates weak role–filler separation or overwritten bindings. \emph{Role swaps} between premises and conclusions point to deficiencies in positional/permutation encoding. More subtle inconsistency across related queries can reflect interference between superposed bindings representing different contexts. Mapping such failures to specific aspects of VSA algebra---separation, capacity, permutation stability---yields diagnostic patterns that guide architectural and training interventions. Thus, VSA-likeness metrics and benchmarks not only evaluate models but also illuminate \emph{why} logical and compositional reasoning succeeds or fails in attention-based LLMs.

\section{Open problems and research agenda}
\label{sec:open-problems}

\subsection{Theoretical questions}
\label{subsec:theoretical-questions}

Our VSA-based interpretation of attention raises foundational questions about the algebraic structure of transformer computation. One line of inquiry concerns \emph{conditions for equivalence}: under what assumptions on initialization, training dynamics, and embedding geometry do attention layers implement an algebra closely matching a VSA binding/unbinding system? For instance, can approximate orthogonality of key/query spaces and sparsity of attention weights guarantee a well-formed binding operator with a similarity-based inverse?

A second line examines transformer \emph{expressivity} in terms of \emph{VSA algebraic capacity}. While recent work connects transformers to classes of formal languages~\citep{strobl-etal-2024-formal}, these results rarely appeal to explicit role–filler algebras. A VSA-centric theory would ask what classes of logical transformations or proof procedures can be simulated with finite role vectors, binding operators, and superposition capacity, and how these capabilities scale with model depth, width, and number of heads. Understanding this capacity may clarify generalization patterns in reasoning tasks.

\subsection{Architectural questions}
\label{subsec:architectural-questions}

Architecturally, a key question is the \emph{granularity} at which VSA structure should be imposed. Should binding and superposition be enforced at the token level (via specialized embeddings or heads), at the layer level (dedicated binding layers), or via a separate hyperdimensional memory module accessed through cross-attention~\citep{graves2016dnc,frady2021variablebinding}? Each option trades off interpretability, capacity, and computational overhead.

Another question concerns balancing \emph{VSA-like rigidity} with \emph{neural flexibility}. Strict VSA constraints (fixed binding operators, hard orthogonality) may improve symbolic behavior but limit exploitation of statistical patterns in natural data. Fully unconstrained architectures, however, risk losing necessary binding discipline. Hybrid designs---with some heads or modules explicitly VSA-like and others free-form---may reveal how training dynamics distribute symbolic versus statistical responsibilities.

\subsection{Neuro-symbolic and tool-use integration}
\label{subsec:nesy-tool-integration}

A further agenda concerns integration of VSA-style internal states with \emph{external symbolic systems}. If transformers maintain hyperdimensional representations of formulas, partial proofs, or knowledge-graph fragments, how should these interface with logic and theorem provers, SMT solvers, or probabilistic programs~\citep{serafini2016ltn,manhaeve2018deepproblog,creswell2022selectioninference,mialon2023augmented}? One possibility is to design encoders mapping symbolic structures into VSA representations compatible with the model's internal algebra, and decoders transforming hyperdimensional states back into human-readable proofs or explanations. Similar considerations apply to ontologies and knowledge graphs~\citep{nickel2016kgEmbeddingReview}. VSA encodings may bridge graph-based reasoning and text-based LLM reasoning, while multi-agent or tool-augmented systems could use VSA memory as a shared workspace for exchanging structured information. Determining how to design, train, and coordinate such shared hyperdimensional stores remains an open challenge.

\subsection{Bridging to cognitive science and neuroscience}
\label{subsec:cognitive-bridge}

Finally, the VSA interpretation invites interdisciplinary work on the \emph{cognitive plausibility} of binding in LLMs. VSAs have roots in theories of human working memory and variable binding~\citep{smolensky1990tensor,plate1995hrR,kanerva2009hyperdimensional}. If attention approximates a vectorial binding/unbinding algebra, how closely does this mirror hypothesized neural mechanisms for symbolic reasoning and language~\citep{marcus2001algebraicMind}? Are there biological correlates of role–filler separation and structured superposition, and can LLM error patterns (e.g., variable confusion, scope errors) inform cognitive models?

Concrete directions include comparing human-like reasoning errors to VSA failure modes, testing whether increasing VSA-likeness produces more human-like reasoning trajectories, and designing neurocognitive experiments probing for VSA-like representations. Progress here could enrich both AI and cognitive science, and inspire architectures grounded in principles of human symbolic computation.

\section{Conclusion}
\label{sec:conclusion}

Transformer attention can be interpreted as a \emph{soft vector-symbolic algebra} in which queries and keys define role subspaces, values encode fillers, and attention weights implement a differentiable unbinding operator that retrieves and recombines fillers according to role similarity. Residual connections act as superposition, enabling layered compositions of bound structures~\citep{vaswani2017attention,plate1995hrR,kanerva2009hyperdimensional,Kleyko2022vsa}. This viewpoint unifies many empirical findings on LLM behavior, providing an algebraic explanation for how transformers sometimes succeed at symbolic manipulation and why they often fail when interference, entanglement, or dense attention patterns disrupt clean role–filler structure.

Interpreting attention through the VSA lens yields concrete implications for logical reasoning: transformer performance improves when role–filler separation is stable, superposition is controlled, and binding/unbinding behavior approximates principled algebraic operations; it degrades when these properties break down, producing inconsistency, brittleness, and poor systematic generalization~\citep{keysers2020measuring,creswell2022selectioninference}. This perspective motivates new directions for theory, architecture, and evaluation, ranging from explicit binding heads and hyperdimensional memory layers to VSA-likeness metrics and logic-sensitive benchmarks. By framing reasoning in LLMs as approximate computation in a high-dimensional symbolic algebra, this review aims to provide a foundation for designing more interpretable, reliable, and compositionally robust models.

\section{Broader Impact}

This work does not raise ethical or societal concerns. It advances theoretical understanding of transformer-based language models by interpreting attention and residual streams through the algebraic framework of Vector Symbolic Architectures. These insights may help guide the development of more reliable and interpretable reasoning systems, potentially improving the robustness of future machine learning models without altering their societal or ethical risk profile.

\bibliography{anonymous-submission-latex-2026}

% Check whether the conference requires a reproducibility checklist to be included in the paper.
% If so, you can uncomment the following line and ajust the path to include it.
% \input{../../ReproducibilityChecklist/LaTeX/ReproducibilityChecklist.tex}

\end{document}